\documentclass[letterpaper]{article} 
\usepackage{aaai24}
\usepackage{times}  
\usepackage{helvet}  
\usepackage{courier}  
\usepackage[hyphens]{url}  
\usepackage{graphicx} 
\usepackage{natbib}  
\setcitestyle{numbers,square}
\usepackage{caption} 
\usepackage{algorithm}
\usepackage{algorithmic}
\usepackage{float}                  
\usepackage{subfig}
\usepackage{amsmath}
\usepackage{newfloat}
\usepackage{listings}
\DeclareCaptionStyle{ruled}{labelfont=normalfont,labelsep=colon,strut=off} 
\floatstyle{ruled}
\newfloat{listing}{tb}{lst}{}
\floatname{listing}{Listing}
\title{BARET : Balanced Attention based Real image Editing driven by Target-text Inversion}
\author {
    Yuming Qiao\textsuperscript{\rm 1,2},
    Fanyi Wang\textsuperscript{\rm 1},
    Jingwen Su\textsuperscript{\rm 1},
    Yanhao Zhang\textsuperscript{\rm 1},
    Yunjie Yu\textsuperscript{\rm 1},
    Siyu Wu\textsuperscript{\rm 3},
    Guo-Jun Qi\textsuperscript{\rm 1,4}
}
\affiliations {
    \textsuperscript{\rm 1}OPPO Research Institute\\
    \textsuperscript{\rm 2}Tsinghua University\\
    \textsuperscript{\rm 3}Zhejiang University\\
    \textsuperscript{\rm 4}Westlake University\\
    qym21@mails.tsinghua.edu.cn, 
    wangfanyi@oppo.com, 
    j\_su95@outlook.com,
    zhangyanhao@oppo.com,
    yunjieyu007@gmail.com,
    22151365@zju.edu.cn,
    guojunq@gmail.com
}

\begin{document}
\twocolumn[{%
\renewcommand\twocolumn[1][]{#1}%
\maketitle
\begin{center}
    \captionsetup{type=figure}
    \includegraphics[width=\linewidth]{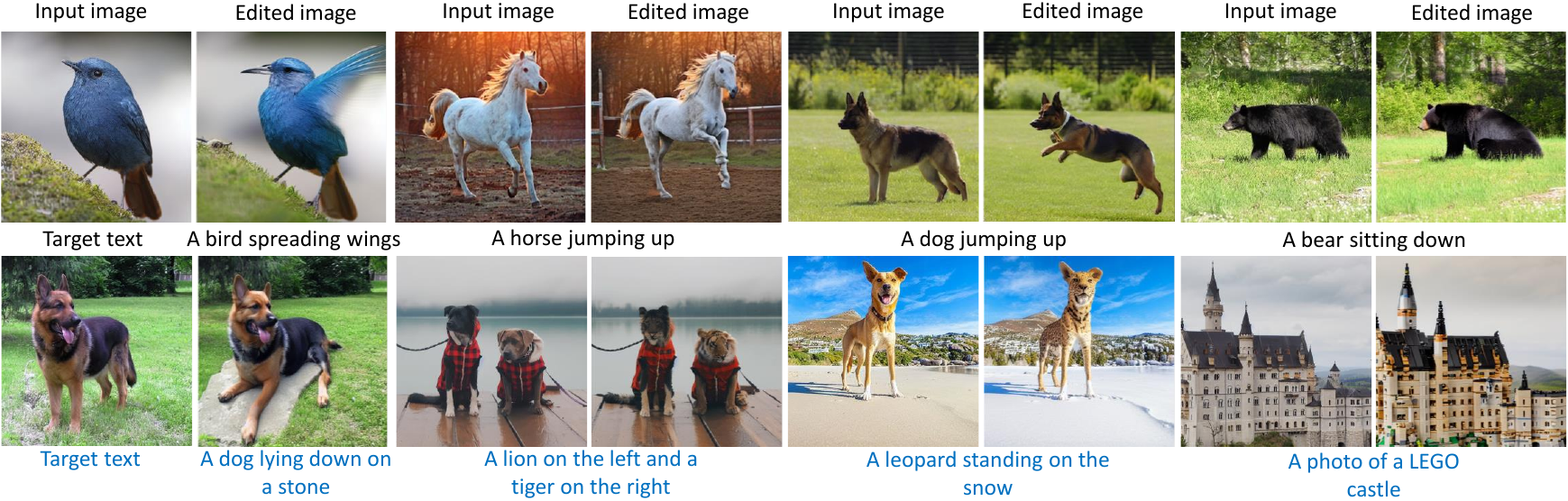}
    \captionof{figure}{\textbf{BARET Real Image Editing.} Our method supports various editing types like style transfer, background and foreground editing, and complex editing with non-rigrid changes such as posture changes. Here, we show some pairs of input real images and editing results aligned with target texts. }
    \label{fig:Firstshow}
\end{center}%
}]

\begin{abstract}

Image editing approaches with diffusion models have been rapidly developed, yet their applicability are subject to requirements such as specific editing types (e.g., foreground or background object editing, style transfer), multiple conditions (e.g., mask, sketch, caption), and time consuming fine-tuning of diffusion models. For alleviating these limitations and realizing efficient real image editing, we propose a novel editing technique that only requires an input image and target text for various editing types including non-rigid edits without fine-tuning diffusion model. Our method contains three novelties:
(I) Target-text Inversion Schedule (TTIS) is designed to fine-tune the input target text embedding to achieve fast image reconstruction without image caption and acceleration of convergence.
(II) Progressive Transition Scheme applies progressive linear interpolation between target text embedding and its fine-tuned version to generate transition embedding for maintaining non-rigid editing capability.
(III) Balanced Attention Module (BAM) balances the tradeoff between textual description and image semantics.
By the means of combining self-attention map from reconstruction process and cross-attention map from transition process,
the guidance of target text embeddings in diffusion process is optimized.
In order to demonstrate editing capability, effectiveness and efficiency of the proposed BARET, we have conducted extensive qualitative and quantitative experiments. Moreover, results derived from user study and ablation study further prove the superiority over other methods.


\end{abstract}

\section{Introduction}
Recently, large Text-to-Image (T2I) models \citep{ramesh2022hierarchical, saharia2022imagen, rombach2022SD} have demonstrated powerful generation capabilities that have drawn a great deal of attentions and attracted interests from academia and industry.
Among image generation tasks, real image editing has been an interesting and valuable topic for a long time.
Real image editing refers to the controllable editing of a real image guided by specific condition (e.g., text, mask, sketch and additional views of the object) \citep{mout2iadapter, zhang2023controlNet}, while preserving the original image features as much as possible. In these numerous conditions, textual condition is the most concise and intuitive. Hence text-guided real image editing has become one of the most popular research topics.

Previous leading methods on text-guided image editing have shortcomings in different aspects. 
(I) limited editing capabilities such as object editing, background/foreground editing, and style transfer, etc.\citep{OmriBlended, DiffusionCLIP, bar2022text2live}, while incapable of non-rigid image editing. 
(II) image captions are required for later modification, but make the editing method inconvenient to be applied \citep{mokady2023null-text}. 
(III) fine-tuning the diffusion model on a single image leads to risks of undermining the pretrained model \citep{kawar2023imagic, zhang2023sine}.
To combat these deficiencies, we propose a novel text-based real image editing method named BARET. Only the original image and a simple target text of editing are needed to realize highly controllable image editing, including both basic editing scenarios and complex edits (i.e. non-rigid changes).
As shown in the lower left corner of Fig. \ref{fig:Firstshow}, when a simple text prompt ``a dog lying down on a stone" is provided, the proposed BARET can realize posture change from standing to lying down on a stone of the same dog in the original image.

In this paper, the proposed BARET consists of three components. 
(I) Target-text inversion schedule circumvents the requirement of image caption and improves the efficiency of original image content extraction. Based on DDIM inversion, target text embedding is iteratively fine-tuned in DDIM sampling process for image reconstruction.
(II) Progressive transition scheme enhances editing capabilities especially non-rigid edits. It integrates the target text information and original image content in the generated transition result, by applying progressive linear interpolation between fine-tuned embedding and target text embedding. 
(III) Balanced Attention Module is proposed to interact the self-attention map of reconstruction process, the cross attention map of transition process and the target embedding in the editing process. While maintaining the original image layout features, the capability of non-rigid editing is improved by combination of non-rigid semantic information from transition process and target text embedding.
We have conducted a user study, based on a series of comparison experiments with leading text-guided image editing methods. Results have shown that our BARET is superior in both objective metrics and user preference, especially on the task of non-rigid edits. Furthermore, ablation study on the hyperparameters of interpolation parameter, injection steps of self-attention map and cross attention map, has been conducted as well.
Our contributions in this paper are summarised as follows,
\begin{itemize}
    \item Regarding of the additional image captions, we propose target-text inversion schedule to fine-tune target text embeddings for image reconstruction. It makes the editing process more concise, takes only 16s on a single A100, and avoids information loss during fine-tuning diffusion model.
    \item On the subject of the complex non-rigid edits, we propose a progressive transition scheme to transform non-rigid textual information to image semantics of transition result. 
    \item In respect of balancing original image information and non-rigid change information, we further propose Balance Attention Module (BAM). BAM enhances the complex non-rigid editing capability, by manipulating the self-attention map of the reconstruction process and the cross-attention map of the transition process.
\end{itemize}

\section{Related Work}
Large-scale text-to-image diffusion models \citep{saharia2022imagen,rombach2022SD,chang2023muse,gu2022vector,song2023consistency,ramesh2022hierarchical, song2019generative} significantly improve the quality of generated images under textual conditions. 
Nevertheless, it is still challenging to achieve specific editing on real images, because such task is rarely realized by means of training. Mask guided diffusion models that denoise only in the masked region become popular for real image editing \citep{rombach2022SD,Nichol2022GLIDE,xie2023smartbrush}.
However the interactive experience of mask-based editing is not quite user-friendly, and the limited generated content due to the mask might lead to distortion for object editing. Hence, text based image editing methods are proposed successively. Liu et al.\citep{Liu2023DiffCLIP} proposed semantic diffusion guidance to incorporate information of the reference text and the image in the diffusion process.
Hertz et al. \citep{HERtz2023P2P} proposed to utilize the cross-attention map between text and image in the diffusion process for controllable real image editing. Inspired by the GAN inversion strategy \citep{yeh2017semantic,zhu2016generative,xia2022gan,lipton2017precise,creswell2018inverting,bermano2022state}, null-text inversion \citep{mokady2023null-text} method extracts image caption by manual or off-the-shelve captioning model \citep{mokady2021clipcap} in advance, and fine-tunes unconditional embedding for image reconstruction, then edits the reconstructed image by prompt2prompt \citep{HERtz2023P2P}. 
These methods do not require fine-tuning diffusion model, but are limited in editing performance in specific contexts. 

Tuning model parameters during inversion \citep{roich2022pivotal,tov2021designing,valevski2022unitune} is also applied to the diffusion model based image editing task. Dreambooth \citep{ruiz2023dreambooth} fine-tunes the whole UNet with a few images of a specific subject and a unique identifier with class-specific prior preservation loss to achieve subject-driven generation. Kawar et al \citep{kawar2023imagic} proposed Imagic that achieves image reconstruction by fine-tuning target text embedding and diffusion model. And image is edited by denoising process with interpolated embedding between target text embedding and fine-tuned embedding.
SINE \citep{zhang2023sine} proposed model-based guidance strategy that combines the pre-trained diffusion model and its fine-tuned version to edit input image.

We fully consider characteristics of existing text-only image editing methods and propose a novel image editing method named BARET. Our method has high capability of non-rigid edits, and does not require an additional image caption or fine-tuning diffusion model, thus significantly reduces inference time.

\section{Methodology}
Given the real image $I_{org}$, our goal is to solely rely on the target text $X_{tgt}$ for editing to derive the result $I_{edt}$ that preserves most of the original image features. To achieve this goal, a novel method named BARET is proposed in this paper to address shortcomings of previous methods.
BARET contains three components, which are Target-Text Inversion Schedule, Progressive Transition Scheme and Balanced Attention Module. 
\begin{figure}[t]
    \centering
    \includegraphics[width=\linewidth]{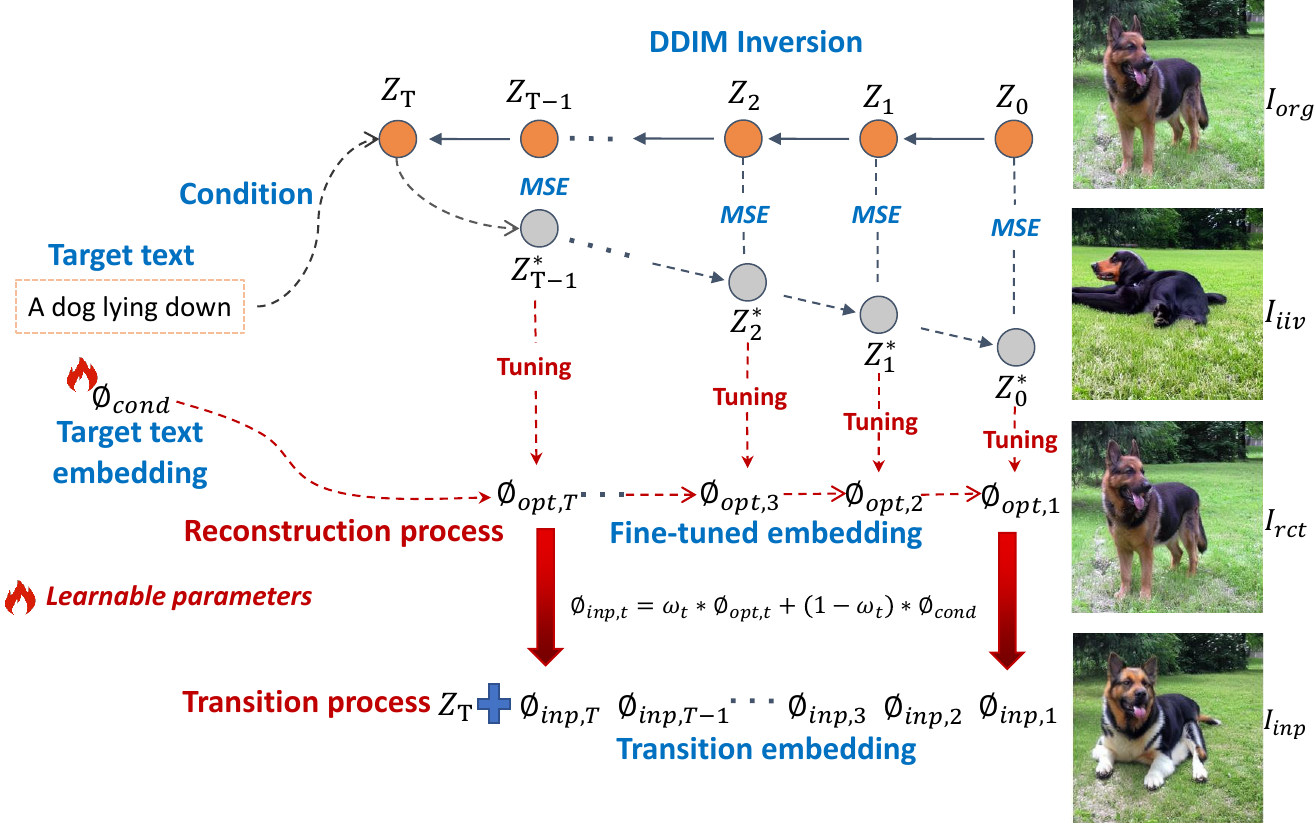}
    \caption{Illustration of our Target-text Inversion Schedule and Progressive transition scheme.}
    \label{fig:TTIS}
\end{figure}

\begin{algorithm}[tb]
\caption{Target-text inversion}
\label{alg:algorithm}
\textbf{Input}: Target text ${{X}_{tgt}}$, input image ${{I}_{org}}$ \\
\textbf{Output}: Fine-tuned embedding $[{{\phi}_{opt,T}},...,{{\phi}_{opt,1}}]$ \\
\begin{algorithmic}[1] 
\STATE{Get ${{Z}_{0}}\sim{{Z}_{T}}$ from ${{I}_{org}}$ via DDIM inversion with guidance scale = 1. Get ${{Z}_{T-1}^{*}}\sim{{Z}_{0}^{*}}$ via Initial inversion under target text embedding ${{\phi}_{cond}}$ and noisy latent ${{Z}_{T}}$ with guidance scale = 7.5}.
\FOR{$t=T,T-1,...,1$}
\STATE{${{\phi}_{opt,t,0}}=\varphi({{X}_{tgt}})$}
\FOR{$i=1,2,...,N$}
\STATE{${{\phi}_{opt,t,i}}=\lambda(\pounds({{Z}_{t-1}^{*}},{{Z}_{t-1}}),{{\phi}_{opt,t,i-1}})$}
\STATE{${{{Z}_{t-1}^{*}}}={{\varepsilon}_{\theta}}({{Z}_{t}^{*}},t,{{\phi}_{opt,t}})$}
\ENDFOR
\STATE{${{\phi}_{opt,t}}={{\phi}_{opt,t,N}}$}
\ENDFOR
\STATE\textbf{Return}: {$[{{\phi}_{opt,T}},{{\phi}_{opt,T-1}},...,{{\phi}_{opt,1}}]$}
\end{algorithmic}
\end{algorithm}
\subsection{Target-Text Inversion Schedule}
Considering simplicity and efficiency of editing, this paper proposes target-text inversion schedule. The advantage is that there is no need to prepare the image caption in advance, and it greatly reduces fine-tuning time during reconstruction process. 
With the premise that only target text is given, fine-tuning null text embedding requires longer training time to realize image reconstruction. Since generation result in the classifier free guidance strategy is largely guided by the input text, when there exists mismatch between the target text and the image, more adjustments on null text embedding is necessary. On the contrary, direct fine-tuning the target text embedding can realize image reconstruction with fewer steps and faster convergence. Comparison of inference time of different methods is provided in Table \ref{Table:infertimecomp}. Target text inversion process is shown in Algorithm 1 and Fig. \ref{fig:TTIS}. Firstly, DDIM inversion applied on  ${{I}_{org}}$ derives initial diffusion trajectory ${{Z}_{0}}\sim{{Z}_{T}}$. Then ${{Z}_{T}}$ and the target text embedding ${{\phi}_{cond}}$ are used as the input to perform conditional DDIM sampling process and fine-tune ${{\phi}_{cond}}$. The fine-tuning process is expressed in the following Equation \ref{con:TTIS_1} and \ref{con:TTIS_2},
\begin{equation}
{{\phi}_{opt,t}}={\lambda(\pounds({{Z}_{t-1}^{*}},{{Z}_{t-1}}),{{\phi}_{cond}})}\label{con:TTIS_1}
\end{equation}
\begin{equation}
{{{Z}_{t-1}^{*}}}={{\varepsilon}_{\theta}}({{Z}_{t}^{*}},t,{{\phi}_{opt,t}})\label{con:TTIS_2}
\end{equation}
where ${{\phi}_{opt,t}}$ is the fine-tuned embedding at the $t$th timestamp, $\lambda(\bullet)$ is the optimizer, $\pounds(\bullet)$ is the reconstruction loss, ${{Z}_{t-1}}$ is the diffusion trajectory of DDIM inversion, ${{Z}_{t-1}^{*}}$ is the intermediate latent code of the DDIM sampling process with ${{Z}_{t}^{*}}$ and fine-tuned embedding ${{\phi}_{opt,t}}$ as the input, ${{\phi}_ {cond}}$ is the target text embedding and also the parameter to be optimized. The overall fine-tuning process takes ${{Z}_{T}}$ and ${{\phi}_{cond}}$ with guidance scale 7.5 as the input, and output ${{Z}_{T-1}^{*}}$. Reconstruction loss between ${{Z}_{T-1}^{*}}$ and ${{Z}_{T-1}}$ is used to optimize ${{\phi}_{cond}}$ and obtain ${{\phi}_{opt,T}}$. Fine-tuned ${{\phi}_{opt,T}}$ and ${{Z}_{T}}$ are used to update ${{Z}_{T-1}^{*}}$ to fine-tune iteratively, which finally brings sampling process ${\{{Z}_{t}^{*}\}}_{t=0}^{T}$ closer to initial diffusion trajectory ${\{{Z}_{t}\}}_{t=0}^{T}$.




\subsection{Balanced Attention Module}
\begin{figure}[htbp]
    \centering
    \includegraphics[width=\linewidth]{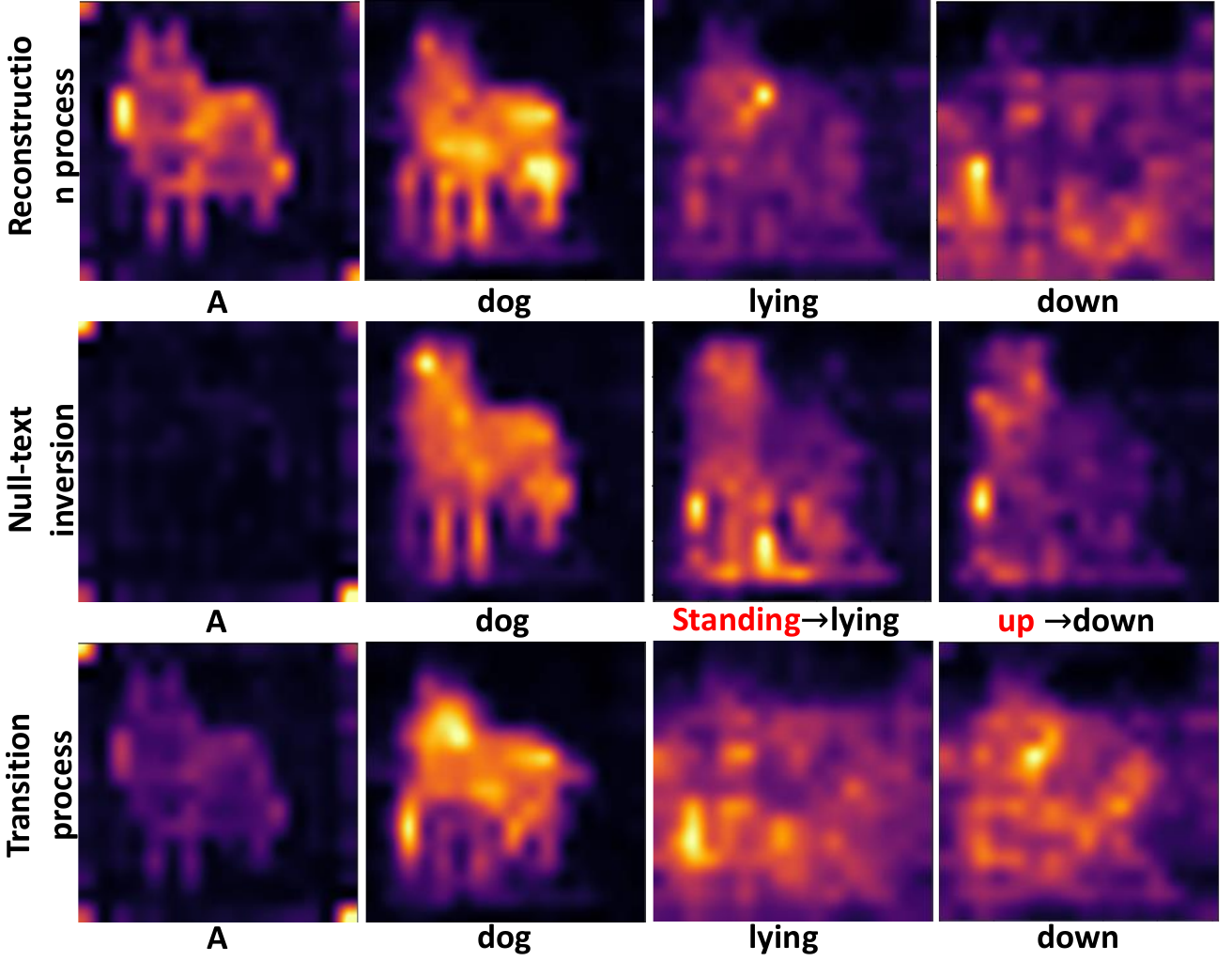}
    \caption{Illustration of the effectiveness of our transition process in non-rigid edits.}
    \label{fig:attnmap}
\end{figure}

Existing attention-based methods directly manipulate self and cross attention maps of fine-tuned embedding. These methods often fail to achieve the expected results for edits with non-rigid changes. Although fine-tuned embedding has constructed strong spatial feature correlation with the original image, it is unable to produce the structural changes required for editing in the attention map. Hence the expected non-rigid changes cannot be manifested in the generated results. For example, taking ``a dog lying down" and Fig. \ref{fig:Firstshow} bottom-left image as inputs, correlations between the fine-tuned embedding and the original image are constructed as shown in the first row of Fig. \ref{fig:attnmap}. It is easy to find that the attention map of the reconstructed image does not have the spatial layout information corresponding to ``lying down". And from the corresponding attention map  of ``dog", it can be seen that it is still in standing posture. Second row of Fig. \ref{fig:attnmap} shows the attention map of null-text inversion taking the caption ``a dog standing up" as input for reconstruction. Likewise, attention maps corresponding to the words ``dog" and ``standing" can only reflect the standing posture of the dog. If ``standing up" is replaced by ``lying down" by the means of prompt2prompt \citep{HERtz2023P2P}, it causes conflicts between the textual description of ``a dog lying down" and the semantic information of the dog standing up in the attention map (second row of Fig. \ref{fig:attnmap}). Thus it is difficult to effectively inject non-rigid change information carried by the target text into the sampling process, resulting in unsatisfactory editing results that are not aligned with the given textual description.

\subsubsection{Progressive Transition Scheme.} 
In order to solve the above problems, we propose a novel embedding interpolation method to embody the non-rigid change information from the target text into the generated results. The interpolation method is expressed by the following Equation \ref{con:transition}:
\begin{equation}
{{\phi}_{inp,t}}={{\omega}_{t}}*{{\phi}_{opt,t}}+(1-{{\omega}_{t}})*{{\phi}_{cond}}\label{con:transition}
\end{equation}
where ${{\omega}_{t}}$ and ${{\phi}_{opt,t}}$ are the interpolation parameter and fine-tuned embedding at timestamp $t$ in diffusion process. Since denoising result in early stage largely determines the direction of generation, an interpolation parameter varying from large to small is applied.
A large value assigned to the interpolation parameter at the beginning is helpful to preserve features of the original image. While small values set in the middle and later stages can drive the transition embedding ${{\phi}_{inp,t}}$ closer to the target text embedding ${{\phi}_{cond}}$, injecting the non-rigid change information of the target text.
As shown in Fig. \ref{fig:TTIS}, ${{I}_{rct}}$ is the image reconstruction result generated by fine-tuned embedding ${\{{\phi}_{opt,t}\}_{t=1}^{T}}$ through DDIM sampling, ${{I}_{iiv}}$ is initial inversion result by target text embedding ${{\phi}_{cond}}$, ${{I}_{inp}}$ is transition result sampled with transition embedding ${\{{\phi}_{inp,t}\}_{t=1}^{T}}$ using ${{\omega}_{t}}\sim{U}(0.8,0.1)$.
It can be seen that the initial inversion result represented by ${{I}_{iiv}}$ loses a number of features of both the editing object and the background. The transition result ${{I}_{inp}}$ fails to achieve the editing goal, but it can introduce the target text semantic information while retaining the original image layout and editing object features. As shown in Fig. \ref{fig:attnmap} bottom row, the cross attention map formed between $\{{\phi}_{inp,t}\}_{t=1}^{T}$ and ${{I}_{inp}}$ has obvious non-rigid change information compared with that of the null-text inversion in middle row. Compared to null-text inversion, our transition process can better characterize the non-rigid change semantic information of the target text during the diffusion process. This provides important support for the following Balanced Attention Module.

\begin{figure}[htbp]
    \centering
    \includegraphics[width=\linewidth]{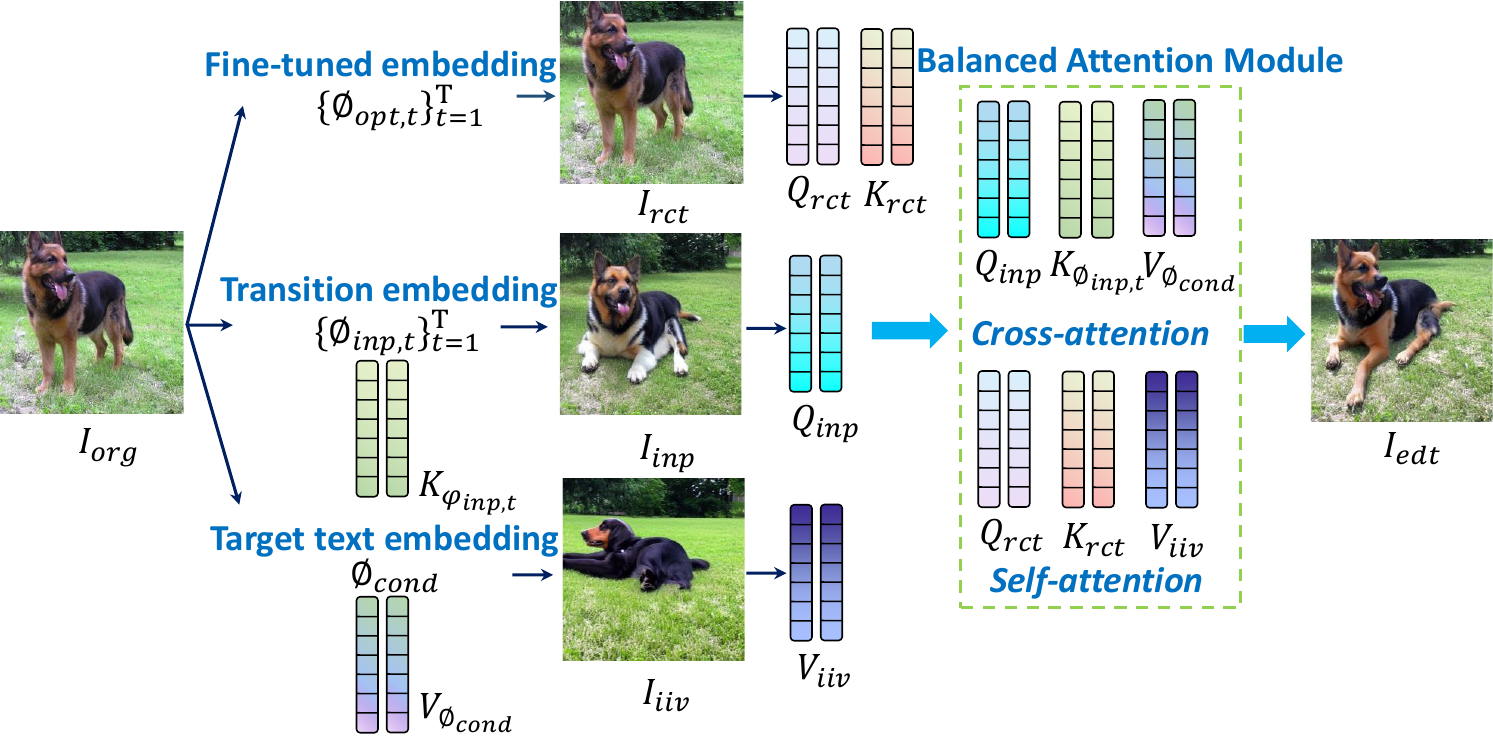}
    \caption{Illustration of Balanced Attention Module.} 
    \label{fig:BAM}
\end{figure}
\subsubsection{Balanced Attention Module.} 
As shown in Fig. \ref{fig:BAM}, we define DDIM process under ${\{{\phi}_{opt,t}\}_{t=1}^{T}}$ as the reconstruction process, DDIM under $\{{\phi}_{inp,t}\}_{t=1}^{T}$ as the transition process, and DDIM under $\phi_{cond}$ as the editing process. 

To retain sufficient original image features in the editing process, in self-attention module original ${{Q}_{iiv}}$ and ${{K}_{iiv}}$ are replaced with ${{Q}_{rct}}$ and ${{K}_{rct}}$ in the reconstruction process. 
To better utilize the non-rigid change information in the transition process, in cross-attention module original ${{Q}_{iiv}}$ and ${{K}_{{\phi}_{cond}}}$ are replaced by ${{Q}_{inp}}$ and ${{K}_{{\phi}_{inp}}}$ from transition process. It is noted that non-rigid semantic of ${{Q}_{inp}}\cdot{{K}_{{\phi}_{inp}}^{T}}$ has a good alignment with non-rigid textual information of ${\phi}_{cond}$, which further improves the non-rigid editing capability.
Modifications on self-attention and cross-attention modules during editing process can be expressed in the following Equations \ref{con:BAMS} and \ref{con:BAMC},
\begin{equation}
\begin{split}
\small
{softmax(\frac{{{Q}_{rct}}\cdot{{K}_{rct}^{T}}}{\sqrt{d}}) \gets softmax(\frac{{{Q}_{iiv}}\cdot{{K}_{iiv}^{T}}}{\sqrt{d}})}\label{con:BAMS}
\end{split}
\end{equation}
\begin{equation}
\begin{split}
\small
softmax(\frac{{{Q}_{inp}}\cdot{{K}_{{\phi}_{inp}}^{T}}}{\sqrt{d}}) \mathbf{\gets} softmax(\frac{{{Q}_{iiv}}\cdot{{K}_{{\phi}_{cond}}^{T}}}{\sqrt{d}})\label{con:BAMC}
\end{split}
\end{equation}

\begin{figure}[htbp]
    \centering
    \includegraphics[width=\linewidth]{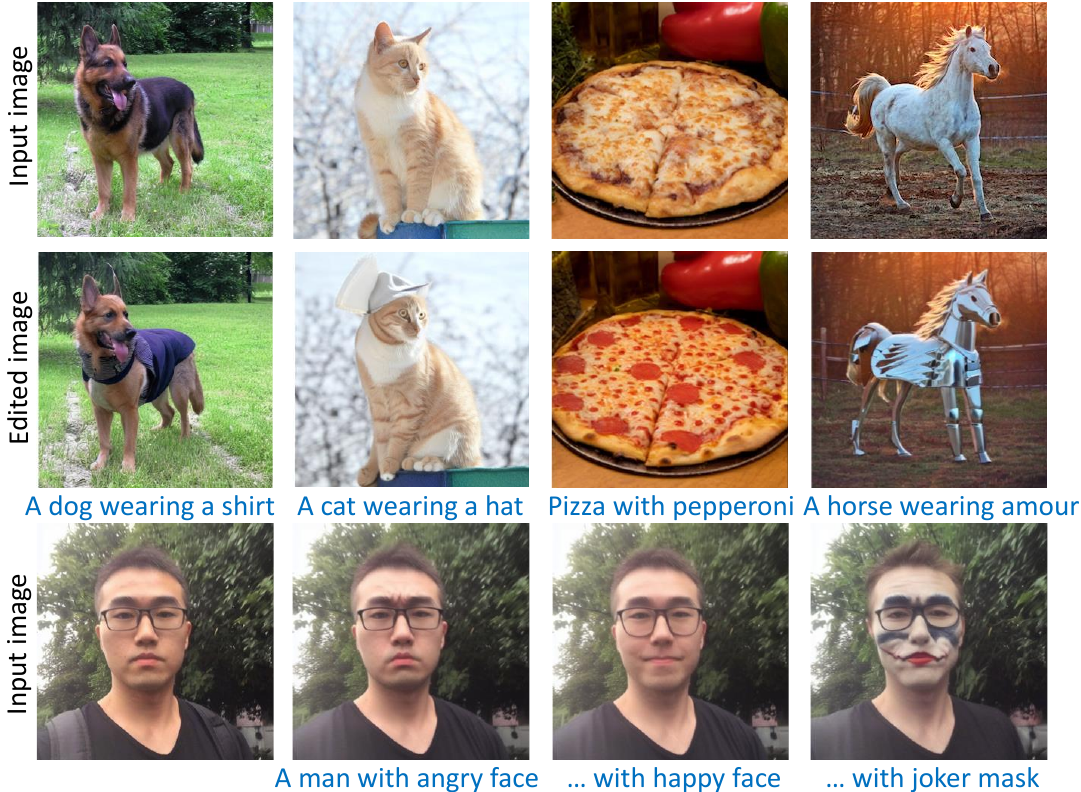}
    \caption{Obejct addition and face manipulation samples of our BARET.}
    \label{fig:faceandadd}
\end{figure}

For editing tasks with rigid changes (e.g., style transfer, foreground and background editing), there is no need to embody the non-rigid change information in the attention map. Correspondingly, interpolation weight ${{\omega}_{t}}$ is set to 1, and transition embedding $\{{\phi}_{inp,t}\}_{t=1}^{T}$ is completely equivalent to the fine-tuned embedding $\{{\phi}_{opt,t}\}_{t=1}^{T}$. Subsequently, BAM introduces self-attention map and cross-attention map from the reconstruction process to the editing process. Hence information of the target text is incorporated on the basis of maintaining most of original image features. As demonstrated in Fig. \ref{fig:Firstshow}, our BARET method has powerful capability of fast arbitrary editing. Additionally, as shown in Fig. \ref{fig:faceandadd}, our BARET also has excellent performance on tasks of object addition and personalized portrait editing.

\section{Experiments}

\subsection{Implementations}

All experiments are based on stable diffusion v1.5 \citep{rombach2022SD}, implemented DDIM sampling strategy with 50 steps and guidance scale 7.5. For the target text inversion, loss function of fine-tuning target text embedding is MSE, tuning iterations are 250 in total and 5 iterations per step, and optimizer is Adam \citep{Kingma2014Adam}. Learning rate is 0.001. Progressive loss value is defined as ${\{{t*1e-5}\}_{t=1}^{T}}$ in each timestamp for early stop, such that reconstruction quality can be boosted with low threshold of loss at the early stage of denoising. And the threshold is gradually raised in the later stage. 
\subsection{Results}
\subsubsection{Comparison.}
\begin{figure*}[t]
    \centering
    \includegraphics[width=\linewidth]{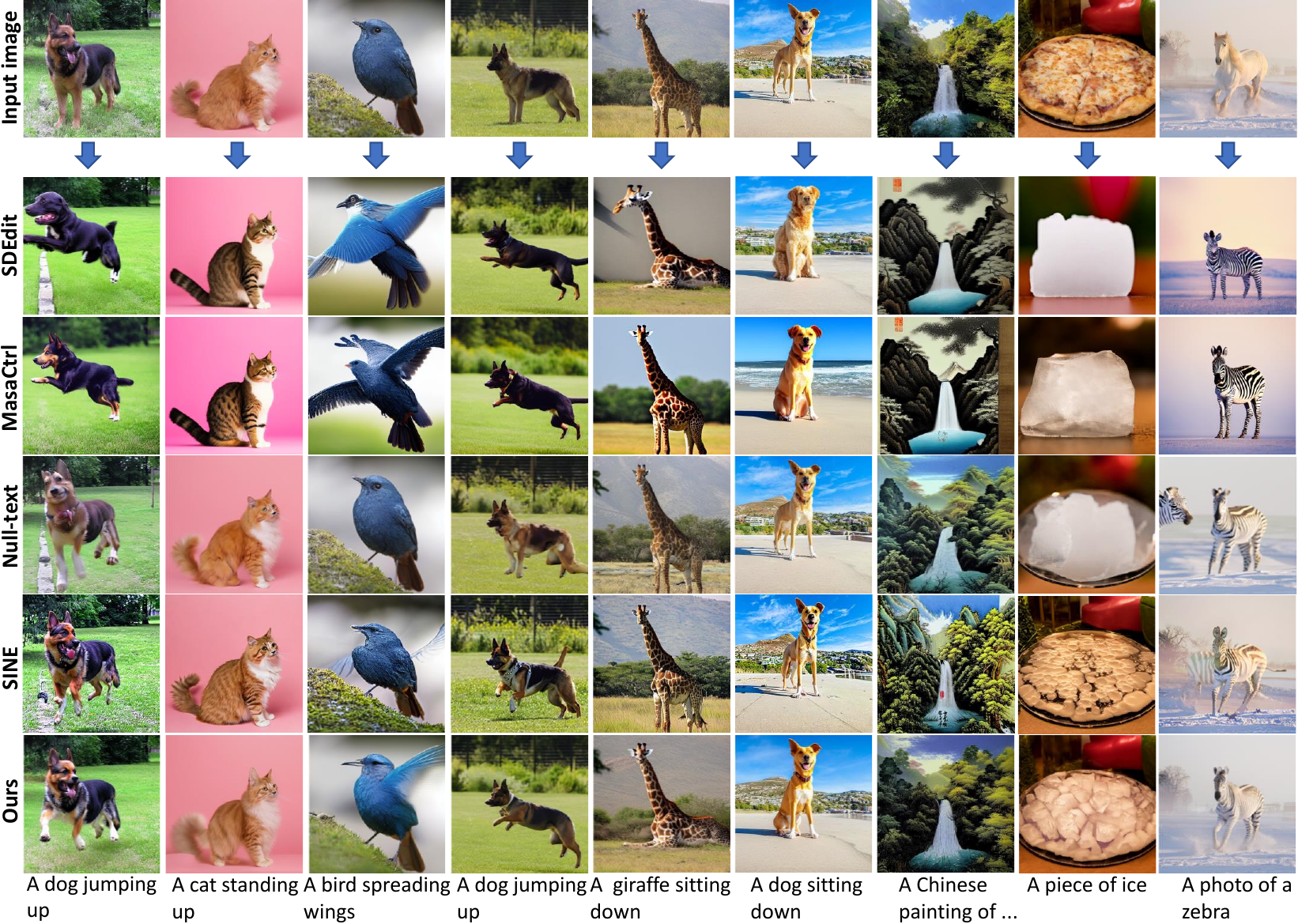}
    \caption{\textbf{Method Comparison.} We compare SDEdite, MasaCtrl, Null-text inversion, and SINE to our method. Our BARET can be applied in arbitrary edits (e.g., posture changes, style transfer, background/object editing, and object addition), while preserving characteristics of original image.}
    \label{fig:compare}
\end{figure*}

We compare our method with the state-of-the-art text-based real image editing methods with code that are publicly available, which are SDEdit \citep{Meng2022SDEdit}, MasaCtrl \citep{cao2023masactrl}, Null-text inversion \citep{mokady2023null-text}, and SINE \citep{zhang2023sine}. 
Their text input format and fine-tune strategy were deployed as described in their papers. It is noted that both Null-text and SINE need prior caption. Fig. \ref{fig:compare} shows the editing results of different methods. The proposed BARET significantly outperforms the other methods, in terms of text alignment between editing results and target text, and image fidelity between editing results and original image, especially on complex non-rigid editing (e.g, bird spreading wings, giraffe lying down, cat standing up and dog sitting down).

\begin{figure}[t]
    \centering
    \includegraphics[width=\linewidth]{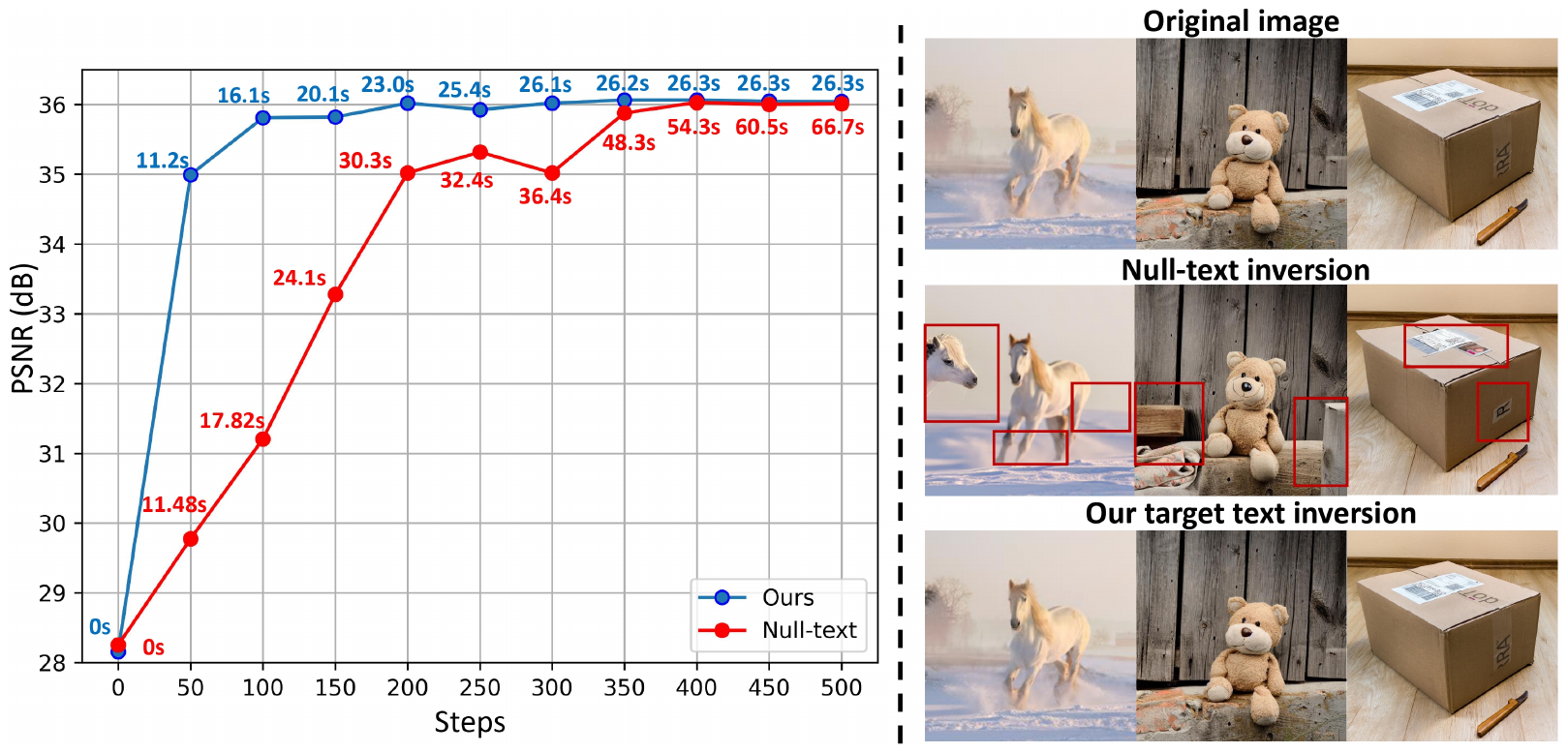}
    \caption{Comparison of our Target-Text Inversion Schedule and Null-text inversion.}
    \label{fig:PSNRV2}
\end{figure}

\begin{table}[h]
\small
\setlength\tabcolsep{3pt}
\centering
{
\begin{tabular}{ccccc}
\hline
Method                                 & Inversion & Editing & Non-rigrid edits & Multiple edits \\
\hline
SDEdite            & /      & 10s    & Yes      & Yes               \\
SINE             & $~$20m      & 10s    & Yes      & Yes               \\
MasaCtrl           & /      & 10s    & Yes      & Yes                \\
Null-text           & $~$1m      & 10s    & No      & Yes               \\
\textbf{Ours}         & \textbf{$~$16s}      & \textbf{10s}       & \textbf{Yes}      & \textbf{Yes}                   \\
\hline      
\end{tabular}
\caption{Inference time comparison.}\label{Table:infertimecomp}}
\end{table}
In addition, our method has obvious advantage on efficiency. The inversion stage of our method takes only about 16s on a single A100, which greatly improves the editing efficiency compared to methods that require fine-tuning diffusion model such as SINE and Imagic. They take longer tuning time around 10$\sim$20 minutes. While the null-text inversion, which is the most efficient among methods for comparison, takes at least about 50s to complete fine-tuning. As shown on the left side of Fig. \ref{fig:PSNRV2}, the proposed TTIS method converges significantly faster than the null-text inversion method, and completes convergence at 100steps in 16.1s, achieving superior reconstruction results. But null-text inversion needs more steps ($>$400 steps) and longer time (48$\sim$66s) to complete. Moreover, null-text inversion has a higher probability of reconstruction failure compared to our TTIS as shown in Fig. \ref{fig:PSNRV2} right column. Because target-text embedding is a better initialization parameter than null-text embedding in the semantic space, which leads to fewer iteration steps to converge and a better reconstruction robustness.  Also, from the aspect of back-propagation, as shown in Equations \ref{con:TTISfast_1}, \ref{con:TTISfast_2} and \ref{con:TTISfast_3}, the classifier guidance scale parameter ${\omega}$ is larger compared to that of fine-tuning null-text embedding, which constitutes one of the reasons that accelerate TTIS.
\begin{equation}
\small
\frac{\partial{{\pounds}({Z}_{t-1},{{{Z}_{t-1}^{*}}})}}{\partial{Y}}\sim\frac{\partial{DM({Z}_{t},{t},{Y},{\oslash})}}{\partial{Y}}\label{con:TTISfast_1}
\end{equation}
\begin{equation}
\small
\frac{\partial{DM({Z}_{t},{t},{Y},{\oslash})}}{\partial{Y}}\sim\nabla{DM({Z}_{t},{t},{Y},{\oslash})}*\frac{\partial{{\varepsilon}_{\theta}}({{Z}_{t}},{t},{Y},{\oslash})}{\partial{Y}}\label{con:TTISfast_2}
\end{equation}
\begin{equation}
\small
\frac{\partial{{\varepsilon}_{\theta}}({{Z}_{t}},{t},{Y},{\oslash})}{\partial{Y}}\sim{\omega}*\frac{\partial{{\varepsilon}_{\theta}}({{Z}_{t}},{t},{Y})}{\partial{Y}}\label{con:TTISfast_3}
\end{equation}

\begin{figure}[t]
    \centering
    \includegraphics[width=\linewidth]{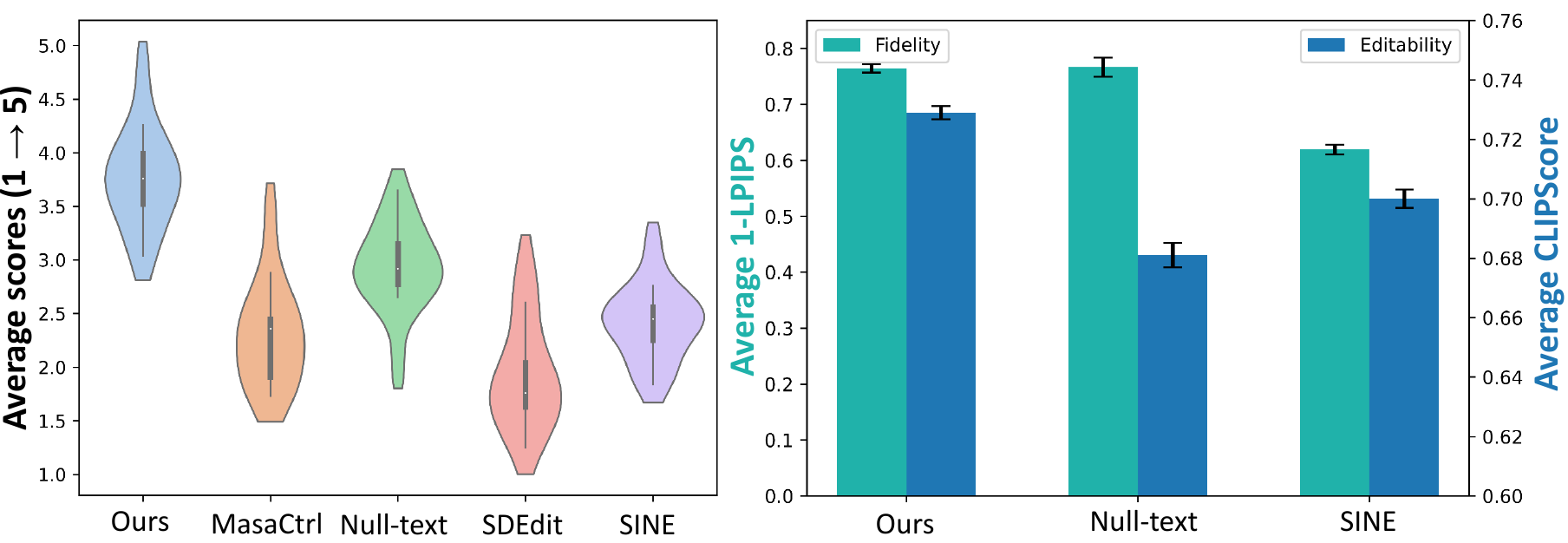}
    \caption{Qualitative analysis of our BARET compared to other leading image editing approaches.}
    \label{fig:UserCLIPLPIPS}
\end{figure}

\subsubsection{User study.} 
Due to the lack of rigorous and accurate quantitative metrics for real image editing tasks, we have conducted a user study to quantify visual effects of various editing methods. To this end, we refer to TEdBench \citep{kawar2023imagic} and collected 100 pairs of real image and textual description for editing. Contents present to the user are 100 pairs of {real image, editing target text, editing result of five methods}.
The five methods are the proposed BARET and four baselines: MasaCtrl, SINE, Null-text, SDEdit. And the scoring process is blind for the user to know which method corresponds to the editing result.  
The user is asked to score the editing results of the five methods ranging from 1 to 5, with 5 representing the best quality that has good alignment with target text and preserves characteristics of original image.
A total of 62 valid results were collected and summarized in Fig. \ref{fig:UserCLIPLPIPS} (left). The average of scores of each method of 100 editing results was computed. We find that our BARET (mean: 3.77, std: 0.20) significantly outperforms the Null-text+p2p (mean: 2.93, std: 0.16), MasaCtrl (mean : 2.31, std: 0.23), SDEdite (mean: 1.93, std: 0.24) and SINE (mean: 2.42, std: 0.12). 

Moreover, the top three ranked methods in user study (i.e., Our BARET, Null-text and SINE) are further evaluated by calculating 1-LIPIS \citep{zhang2018LPIPS} and CLIPScore \citep{ Hessel2021CLIPscore, radford2021CLIPscore} metrics.
LPIPS is the perceptual distance between the edited image and the original image, standing for image fidelity. A large value of 1-LPIPS indicates a closer match to the original image.
CLIPScore computes the cosine similarity between the target text embedding and editing result image embedding, representing the degree of alignment between the editing result and the target text.
The larger the value of CLIPScore is, the closer the editing result is to the description of the target text. As shown in Fig. \ref{fig:UserCLIPLPIPS} (right), our BARET (1-LPIPS: 0.764, CLIPScore: 0.729) has the best objective evaluation result. Null-text method has 1-LPIPS of 0.766, which is slightly higher than that of our BARET, but its CLIPScore (0.681) is absolutely lower than our BARET, indicating poor capability in non-rigid editing. 
And our BARET surpasses SINE (1-LPIPS: 0.619, CLIPScore: 0.7) on each objective metric.

\subsection{Ablation Study}
\begin{figure}[t]
    \centering
    \includegraphics[width=\linewidth]{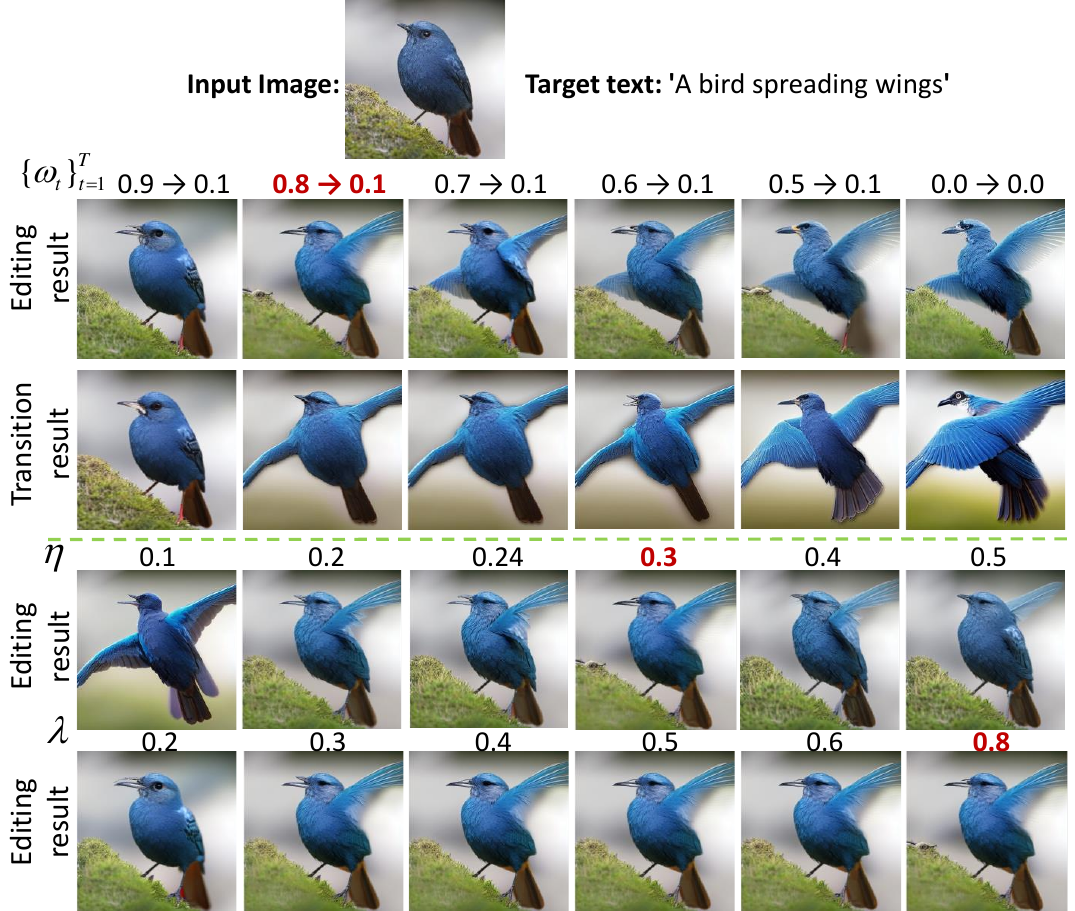}
    \caption{Editing results with various interpolated parameters ${\{{\omega}_{t}\}_{t=1}^{T}}$, self attention injection steps ${\eta}$ and cross attention injection steps ${\lambda}$.}
    \label{fig:Ablationstudy}
\end{figure}

The proposed BARET has three adjustable parameters, interpolation parameter ${\{{\omega}_{t}\}_{t=1}^{T}}$, self-attention injection step ${\eta}$ and cross-attention injection step ${\lambda}$. Interpolation parameter controls the fusion ratio between the non-rigid change information and the original image information in transition process. For example, as shown in the first row of Fig. \ref{fig:Ablationstudy}, when the initial interpolation parameter is large (e.g., 0.9-0.1), transition process is prone to generate images similar to the original image. As the interpolation parameter decreases (e.g., 0.8-0.1), transition process can better merge the non-rigid change information and the original object features. 
\begin{figure}[t]
    \centering
    \includegraphics[width=\linewidth]{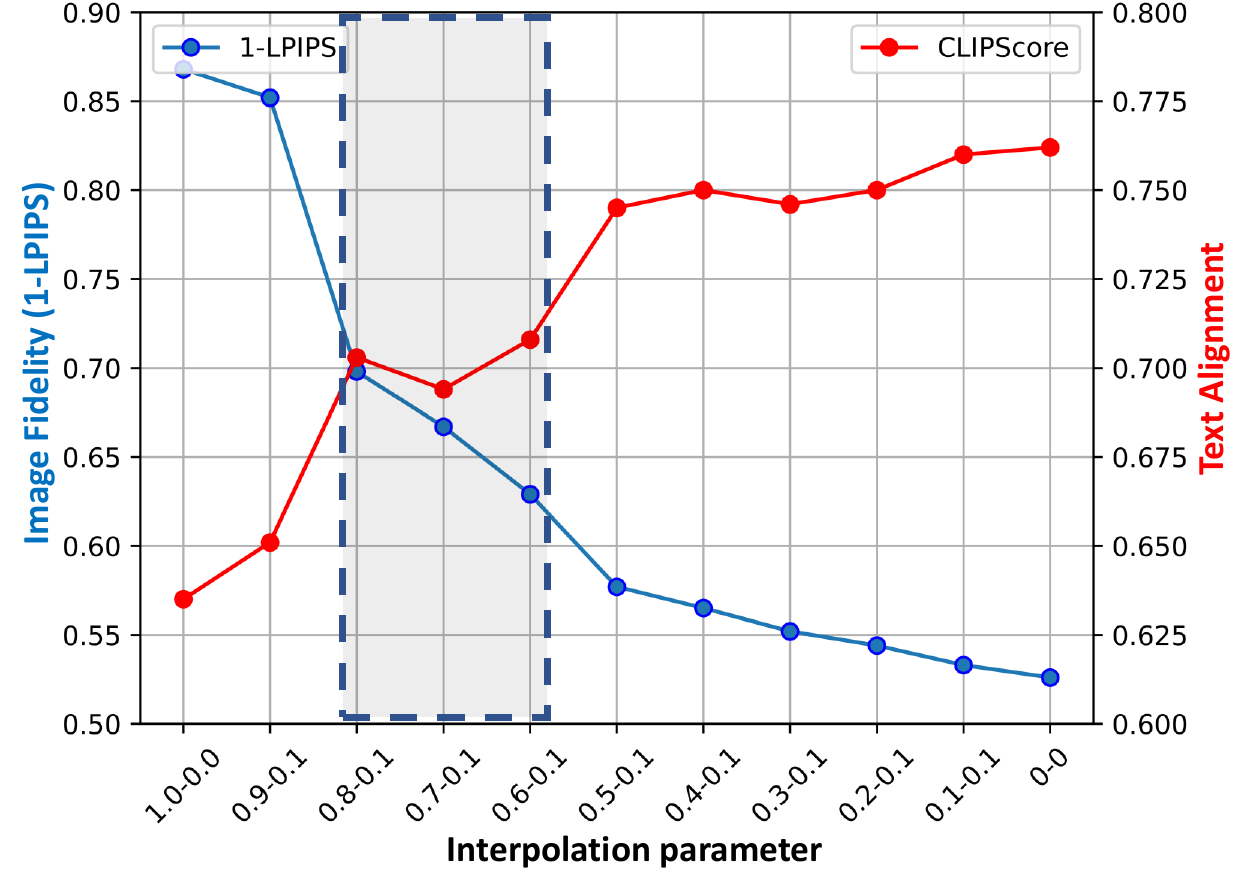}
    \caption{\textbf{Editability–fidelity tradeoff.} CLIP score (target text alignment) and 1-LPIPS (input image fidelity) as functions of interpolation parameter. Edited images tend to match both the input image and target text in the highlighted area.}
    \label{fig:tradeoff}
\end{figure}
From transition results in Fig. \ref{fig:Ablationstudy}, it can be seen that while maintaining characteristics of the original bird, motion information ``spreading wings" is also incorporated. Although most background features are lost,  it can be mitigated by the self-attention maps in the reconstruction process. Correspondingly, it is noticed that editing result under interpolation parameter 0.8-0.1 shows a good tradeoff between characteristics of original image and 
desired non-rigid changes. However, as the interpolation parameter is further reduced, the proportion of the original image feature in the transition result gradually decreases, which is reflected in the editing result that similarity between the editing result and the original image is greatly reduced. As we can observe, when the interpolation parameter is 0.6-0.1, 0.5-0.1, 0-0, the difference of the bird in the editing result and the original image gradually increases.
Similarly, the injection step of self-attention (SA) map and cross-attention (CA) map in BAM is also a very important parameter. As shown in Fig. \ref{fig:Ablationstudy}, when the SA injection step ${\eta}$ is small such as 0.1, the editing result cannot preserve sufficient features of the original image, and has obvious difference to the original image. As ${\eta}$ gradually increases to 0.2-0.3, edited results show a good balance of original content and non-rigid changes. However, as ${\eta}$ further increases to 0.4-0.5, edited results gradually fail to incorporate non-rigid changes and become more similar to the original image.
Unlike the SA injection step ${\eta}$, editing results are less sensitive to the CA injection step ${\lambda}$. When ${\lambda}$ is between 0.4 and 0.8, it has almost no effect on the editing results. But if ${\lambda}$ is further reduced it gradually leads to poor incorporation of non-rigid changes in the editing result.

In summary, we can find that editing results are more sensitive to the interpolation parameter, thus we calculate the LPIPS and CLIPScore of the editing result by varying interpolation parameter. Result analysis is present in Fig. \ref{fig:tradeoff}. As the interpolation parameter decreases, interpolated embedding is closer to target text embedding, non-rigid information increases and the original image information decreases. And for corresponding editing result, CLIPScore gradually increases and 1-LPIPS gradually decreases, indicating that the editing result loses more detailed information of the original image and better align with target text. Considering both image fidelity and text alignment, a reasonable range of the interpolation parameter is 0.8-0.1 $\sim$ 0.6-0.1 (highlighted area in Fig. \ref{fig:tradeoff}) by empirical. In the meanwhile, suggested SA injection steps ${\eta}$ and CA injection steps ${\lambda}$ are 0.2-0.4 and 0.4-0.8 respectively.

\section{Conclusions}
In this paper, we propose BARET, a text based real image editing method with higher efficiency supporting for complex non-rigid edits.
BARET only requires the user to provide a single image and target text for editing. Firstly, the proposed TTIS fine-tunes the target text to extract image content by image reconstruction. Then the proposed progressive transition scheme utilises the target text embedding and fine-tuned embedding to integrate the original image information and the non-rigid structural change information of the target text. Finally, the proposed BAM combines the self-attention map in the reconstruction process and the cross-attention map in the transition process to effectively integrate the characteristics of the original image and the non-rigid change information. Compared with other leading text-based methods, our method has no need to acquire image captions beforehand, and can realize not only simple image editing, but also sophisticated human face manipulation and non-rigid editing, exhibiting best target text alignment and image fidelity. In addition, our method does not require fine-tuning diffusion model, which is a significant advantage in terms of controllability and editing efficiency.

\bibliography{aaai24}

\begin{thebibliography}{10}
\providecommand{\url}[1]{#1}
\csname url@samestyle\endcsname
\providecommand{\newblock}{\relax}
\providecommand{\bibinfo}[2]{#2}
\providecommand{\BIBentrySTDinterwordspacing}{\spaceskip=0pt\relax}
\providecommand{\BIBentryALTinterwordstretchfactor}{4}
\providecommand{\BIBentryALTinterwordspacing}{\spaceskip=\fontdimen2\font plus
\BIBentryALTinterwordstretchfactor\fontdimen3\font minus
  \fontdimen4\font\relax}
\providecommand{\BIBforeignlanguage}[2]{{%
\expandafter\ifx\csname l@#1\endcsname\relax
\typeout{** WARNING: IEEEtran.bst: No hyphenation pattern has been}%
\typeout{** loaded for the language `#1'. Using the pattern for}%
\typeout{** the default language instead.}%
\else
\language=\csname l@#1\endcsname
\fi
#2}}
\providecommand{\BIBdecl}{\relax}
\BIBdecl

\bibitem{OmriBlended}
O.~Avrahami, D.~Lischinski, and O.~Fried, ``Blended diffusion for text-driven
  editing of natural images,'' in \emph{2022 IEEE/CVF Conference on Computer
  Vision and Pattern Recognition (CVPR)}, 2022, pp. 18\,187--18\,197.

\bibitem{DiffusionCLIP}
G.~Kim, T.~Kwon, and J.~C. Ye, ``Diffusionclip: Text-guided diffusion models
  for robust image manipulation,'' in \emph{2022 IEEE/CVF Conference on
  Computer Vision and Pattern Recognition (CVPR)}, 2022, pp. 2416--2425.

\bibitem{mout2iadapter}
C.~Mou, X.~Wang, L.~Xie, Y.~Wu, J.~Zhang, Z.~Qi, Y.~Shan, and X.~Qie,
  ``T2i-adapter: Learning adapters to dig out more controllable ability for
  text-to-image diffusion models,'' 2023.

\bibitem{zhang2023controlNet}
L.~Zhang and M.~Agrawala, ``Adding conditional control to text-to-image
  diffusion models,'' 2023.

\bibitem{bar2022text2live}
O.~Bar-Tal, D.~Ofri-Amar, R.~Fridman, Y.~Kasten, and T.~Dekel, ``Text2live:
  Text-driven layered image and video editing,'' in \emph{European conference
  on computer vision}.\hskip 1em plus 0.5em minus 0.4em\relax Springer, 2022,
  pp. 707--723.

\bibitem{mokady2023null-text}
R.~Mokady, A.~Hertz, K.~Aberman, Y.~Pritch, and D.~Cohen-Or, ``Null-text
  inversion for editing real images using guided diffusion models,'' in
  \emph{Proceedings of the IEEE/CVF Conference on Computer Vision and Pattern
  Recognition}, 2023, pp. 6038--6047.

\bibitem{kawar2023imagic}
B.~Kawar, S.~Zada, O.~Lang, O.~Tov, H.~Chang, T.~Dekel, I.~Mosseri, and
  M.~Irani, ``Imagic: Text-based real image editing with diffusion models,'' in
  \emph{Proceedings of the IEEE/CVF Conference on Computer Vision and Pattern
  Recognition}, 2023, pp. 6007--6017.

\bibitem{zhang2023sine}
Z.~Zhang, L.~Han, A.~Ghosh, D.~N. Metaxas, and J.~Ren, ``Sine: Single image
  editing with text-to-image diffusion models,'' in \emph{Proceedings of the
  IEEE/CVF Conference on Computer Vision and Pattern Recognition}, 2023, pp.
  6027--6037.

\bibitem{saharia2022imagen}
C.~Saharia, W.~Chan, S.~Saxena, L.~Li, J.~Whang, E.~L. Denton, K.~Ghasemipour,
  R.~Gontijo~Lopes, B.~Karagol~Ayan, T.~Salimans \emph{et~al.},
  ``Photorealistic text-to-image diffusion models with deep language
  understanding,'' \emph{Advances in Neural Information Processing Systems},
  vol.~35, pp. 36\,479--36\,494, 2022.

\bibitem{rombach2022SD}
R.~Rombach, A.~Blattmann, D.~Lorenz, P.~Esser, and B.~Ommer, ``High-resolution
  image synthesis with latent diffusion models,'' in \emph{Proceedings of the
  IEEE/CVF conference on computer vision and pattern recognition}, 2022, pp.
  10\,684--10\,695.

\bibitem{ramesh2022hierarchical}
A.~Ramesh, P.~Dhariwal, A.~Nichol, C.~Chu, and M.~Chen, ``Hierarchical
  text-conditional image generation with clip latents,'' \emph{arXiv preprint
  arXiv:2204.06125}, 2022.

\bibitem{Choi2021ILVR}
\BIBentryALTinterwordspacing
J.~Choi, S.~Kim, Y.~Jeong, Y.~Gwon, and S.~Yoon, ``{ILVR:} conditioning method
  for denoising diffusion probabilistic models,'' in \emph{2021 {IEEE/CVF}
  International Conference on Computer Vision, {ICCV} 2021, Montreal, QC,
  Canada, October 10-17, 2021}.\hskip 1em plus 0.5em minus 0.4em\relax {IEEE},
  2021, pp. 14\,347--14\,356. [Online]. Available:
  \url{https://doi.org/10.1109/ICCV48922.2021.01410}
\BIBentrySTDinterwordspacing

\bibitem{Meng2022SDEdit}
\BIBentryALTinterwordspacing
C.~Meng, Y.~He, Y.~Song, J.~Song, J.~Wu, J.~Zhu, and S.~Ermon, ``Sdedit: Guided
  image synthesis and editing with stochastic differential equations,'' in
  \emph{The Tenth International Conference on Learning Representations, {ICLR}
  2022, Virtual Event, April 25-29, 2022}.\hskip 1em plus 0.5em minus
  0.4em\relax OpenReview.net, 2022. [Online]. Available:
  \url{https://openreview.net/forum?id=aBsCjcPu\_tE}
\BIBentrySTDinterwordspacing

\bibitem{xie2023smartbrush}
S.~Xie, Z.~Zhang, Z.~Lin, T.~Hinz, and K.~Zhang, ``Smartbrush: Text and shape
  guided object inpainting with diffusion model,'' in \emph{Proceedings of the
  IEEE/CVF Conference on Computer Vision and Pattern Recognition}, 2023, pp.
  22\,428--22\,437.

\bibitem{Nichol2022GLIDE}
\BIBentryALTinterwordspacing
A.~Q. Nichol, P.~Dhariwal, A.~Ramesh, P.~Shyam, P.~Mishkin, B.~McGrew,
  I.~Sutskever, and M.~Chen, ``{GLIDE:} towards photorealistic image generation
  and editing with text-guided diffusion models,'' in \emph{International
  Conference on Machine Learning, {ICML} 2022, 17-23 July 2022, Baltimore,
  Maryland, {USA}}, ser. Proceedings of Machine Learning Research,
  K.~Chaudhuri, S.~Jegelka, L.~Song, C.~Szepesv{\'{a}}ri, G.~Niu, and
  S.~Sabato, Eds., vol. 162.\hskip 1em plus 0.5em minus 0.4em\relax {PMLR},
  2022, pp. 16\,784--16\,804. [Online]. Available:
  \url{https://proceedings.mlr.press/v162/nichol22a.html}
\BIBentrySTDinterwordspacing

\bibitem{HERtz2023P2P}
\BIBentryALTinterwordspacing
A.~Hertz, R.~Mokady, J.~Tenenbaum, K.~Aberman, Y.~Pritch, and D.~Cohen{-}Or,
  ``Prompt-to-prompt image editing with cross-attention control,'' in \emph{The
  Eleventh International Conference on Learning Representations, {ICLR} 2023,
  Kigali, Rwanda, May 1-5, 2023}.\hskip 1em plus 0.5em minus 0.4em\relax
  OpenReview.net, 2023. [Online]. Available:
  \url{https://openreview.net/pdf?id=\_CDixzkzeyb}
\BIBentrySTDinterwordspacing

\bibitem{Liu2023DiffCLIP}
\BIBentryALTinterwordspacing
X.~Liu, D.~H. Park, S.~Azadi, G.~Zhang, A.~Chopikyan, Y.~Hu, H.~Shi,
  A.~Rohrbach, and T.~Darrell, ``More control for free! image synthesis with
  semantic diffusion guidance,'' in \emph{{IEEE/CVF} Winter Conference on
  Applications of Computer Vision, {WACV} 2023, Waikoloa, HI, USA, January 2-7,
  2023}.\hskip 1em plus 0.5em minus 0.4em\relax {IEEE}, 2023, pp. 289--299.
  [Online]. Available: \url{https://doi.org/10.1109/WACV56688.2023.00037}
\BIBentrySTDinterwordspacing

\bibitem{Gal2023TextualInversion}
\BIBentryALTinterwordspacing
R.~Gal, Y.~Alaluf, Y.~Atzmon, O.~Patashnik, A.~H. Bermano, G.~Chechik, and
  D.~Cohen{-}Or, ``An image is worth one word: Personalizing text-to-image
  generation using textual inversion,'' in \emph{The Eleventh International
  Conference on Learning Representations, {ICLR} 2023, Kigali, Rwanda, May 1-5,
  2023}.\hskip 1em plus 0.5em minus 0.4em\relax OpenReview.net, 2023. [Online].
  Available: \url{https://openreview.net/pdf?id=NAQvF08TcyG}
\BIBentrySTDinterwordspacing

\bibitem{cao2023masactrl}
M.~Cao, X.~Wang, Z.~Qi, Y.~Shan, X.~Qie, and Y.~Zheng, ``Masactrl: Tuning-free
  mutual self-attention control for consistent image synthesis and editing,''
  \emph{arXiv preprint arXiv:2304.08465}, 2023.

\bibitem{Hessel2021CLIPscore}
\BIBentryALTinterwordspacing
J.~Hessel, A.~Holtzman, M.~Forbes, R.~L. Bras, and Y.~Choi, ``Clipscore: {A}
  reference-free evaluation metric for image captioning,'' in \emph{Proceedings
  of the 2021 Conference on Empirical Methods in Natural Language Processing,
  {EMNLP} 2021, Virtual Event / Punta Cana, Dominican Republic, 7-11 November,
  2021}, M.~Moens, X.~Huang, L.~Specia, and S.~W. Yih, Eds.\hskip 1em plus
  0.5em minus 0.4em\relax Association for Computational Linguistics, 2021, pp.
  7514--7528. [Online]. Available:
  \url{https://doi.org/10.18653/v1/2021.emnlp-main.595}
\BIBentrySTDinterwordspacing

\bibitem{radford2021CLIPscore}
A.~Radford, J.~W. Kim, C.~Hallacy, A.~Ramesh, G.~Goh, S.~Agarwal, G.~Sastry,
  A.~Askell, P.~Mishkin, J.~Clark \emph{et~al.}, ``Learning transferable visual
  models from natural language supervision,'' in \emph{International conference
  on machine learning}.\hskip 1em plus 0.5em minus 0.4em\relax PMLR, 2021, pp.
  8748--8763.

\bibitem{zhang2018LPIPS}
R.~Zhang, P.~Isola, A.~A. Efros, E.~Shechtman, and O.~Wang, ``The unreasonable
  effectiveness of deep features as a perceptual metric,'' in \emph{Proceedings
  of the IEEE conference on computer vision and pattern recognition}, 2018, pp.
  586--595.

\bibitem{ho2021classifier}
J.~Ho and T.~Salimans, ``Classifier-free diffusion guidance,'' in \emph{NeurIPS
  2021 Workshop on Deep Generative Models and Downstream Applications}, 2021.

\bibitem{song2020denoising}
J.~Song, C.~Meng, and S.~Ermon, ``Denoising diffusion implicit models,'' in
  \emph{International Conference on Learning Representations}, 2020.

\bibitem{chang2023muse}
H.~Chang, H.~Zhang, J.~Barber, A.~Maschinot, J.~Lezama, L.~Jiang, M.-H. Yang,
  K.~Murphy, W.~T. Freeman, M.~Rubinstein \emph{et~al.}, ``Muse: Text-to-image
  generation via masked generative transformers,'' \emph{arXiv preprint
  arXiv:2301.00704}, 2023.

\bibitem{gu2022vector}
S.~Gu, D.~Chen, J.~Bao, F.~Wen, B.~Zhang, D.~Chen, L.~Yuan, and B.~Guo,
  ``Vector quantized diffusion model for text-to-image synthesis,'' in
  \emph{Proceedings of the IEEE/CVF Conference on Computer Vision and Pattern
  Recognition}, 2022, pp. 10\,696--10\,706.

\bibitem{song2023consistency}
Y.~Song, P.~Dhariwal, M.~Chen, and I.~Sutskever, ``Consistency models,'' 2023.

\bibitem{ruiz2023dreambooth}
N.~Ruiz, Y.~Li, V.~Jampani, Y.~Pritch, M.~Rubinstein, and K.~Aberman,
  ``Dreambooth: Fine tuning text-to-image diffusion models for subject-driven
  generation,'' in \emph{Proceedings of the IEEE/CVF Conference on Computer
  Vision and Pattern Recognition}, 2023, pp. 22\,500--22\,510.

\bibitem{Kingma2014Adam}
\BIBentryALTinterwordspacing
D.~P. Kingma and J.~Ba, ``Adam: {A} method for stochastic optimization,'' in
  \emph{3rd International Conference on Learning Representations, {ICLR} 2015,
  San Diego, CA, USA, May 7-9, 2015, Conference Track Proceedings}, Y.~Bengio
  and Y.~LeCun, Eds., 2015. [Online]. Available:
  \url{http://arxiv.org/abs/1412.6980}
\BIBentrySTDinterwordspacing

\bibitem{ho2020denoising}
J.~Ho, A.~Jain, and P.~Abbeel, ``Denoising diffusion probabilistic models,''
  \emph{Advances in neural information processing systems}, vol.~33, pp.
  6840--6851, 2020.

\bibitem{sohl2015deep}
J.~Sohl-Dickstein, E.~Weiss, N.~Maheswaranathan, and S.~Ganguli, ``Deep
  unsupervised learning using nonequilibrium thermodynamics,'' in
  \emph{International conference on machine learning}.\hskip 1em plus 0.5em
  minus 0.4em\relax PMLR, 2015, pp. 2256--2265.

\bibitem{su2022DDIB}
X.~Su, J.~Song, C.~Meng, and S.~Ermon, ``Dual diffusion implicit bridges for
  image-to-image translation,'' in \emph{The Eleventh International Conference
  on Learning Representations}, 2022.

\bibitem{dhariwal2021diffusion}
P.~Dhariwal and A.~Nichol, ``Diffusion models beat gans on image synthesis,''
  \emph{Advances in neural information processing systems}, vol.~34, pp.
  8780--8794, 2021.

\bibitem{mokady2021clipcap}
R.~Mokady, A.~Hertz, and A.~H. Bermano, ``Clipcap: Clip prefix for image
  captioning,'' \emph{arXiv preprint arXiv:2111.09734}, 2021.

\bibitem{song2019generative}
Y.~Song and S.~Ermon, ``Generative modeling by estimating gradients of the data
  distribution,'' \emph{Advances in neural information processing systems},
  vol.~32, 2019.

\bibitem{bermano2022state}
A.~H. Bermano, R.~Gal, Y.~Alaluf, R.~Mokady, Y.~Nitzan, O.~Tov, O.~Patashnik,
  and D.~Cohen-Or, ``State-of-the-art in the architecture, methods and
  applications of stylegan,'' in \emph{Computer Graphics Forum}, vol.~41,
  no.~2.\hskip 1em plus 0.5em minus 0.4em\relax Wiley Online Library, 2022, pp.
  591--611.

\bibitem{creswell2018inverting}
A.~Creswell and A.~A. Bharath, ``Inverting the generator of a generative
  adversarial network,'' \emph{IEEE transactions on neural networks and
  learning systems}, vol.~30, no.~7, pp. 1967--1974, 2018.

\bibitem{lipton2017precise}
Z.~C. Lipton and S.~Tripathi, ``Precise recovery of latent vectors from
  generative adversarial networks,'' \emph{arXiv preprint arXiv:1702.04782},
  2017.

\bibitem{xia2022gan}
W.~Xia, Y.~Zhang, Y.~Yang, J.-H. Xue, B.~Zhou, and M.-H. Yang, ``Gan inversion:
  A survey,'' \emph{IEEE Transactions on Pattern Analysis and Machine
  Intelligence}, vol.~45, no.~3, pp. 3121--3138, 2022.

\bibitem{zhu2016generative}
J.-Y. Zhu, P.~Kr{\"a}henb{\"u}hl, E.~Shechtman, and A.~A. Efros, ``Generative
  visual manipulation on the natural image manifold,'' in \emph{Computer
  Vision--ECCV 2016: 14th European Conference, Amsterdam, The Netherlands,
  October 11-14, 2016, Proceedings, Part V 14}.\hskip 1em plus 0.5em minus
  0.4em\relax Springer, 2016, pp. 597--613.

\bibitem{yeh2017semantic}
R.~A. Yeh, C.~Chen, T.~Yian~Lim, A.~G. Schwing, M.~Hasegawa-Johnson, and M.~N.
  Do, ``Semantic image inpainting with deep generative models,'' in
  \emph{Proceedings of the IEEE conference on computer vision and pattern
  recognition}, 2017, pp. 5485--5493.

\bibitem{valevski2022unitune}
D.~Valevski, M.~Kalman, Y.~Matias, and Y.~Leviathan, ``Unitune: Text-driven
  image editing by fine tuning an image generation model on a single image,''
  \emph{arXiv preprint arXiv:2210.09477}, 2022.

\bibitem{tov2021designing}
O.~Tov, Y.~Alaluf, Y.~Nitzan, O.~Patashnik, and D.~Cohen-Or, ``Designing an
  encoder for stylegan image manipulation,'' \emph{ACM Transactions on Graphics
  (TOG)}, vol.~40, no.~4, pp. 1--14, 2021.

\bibitem{roich2022pivotal}
D.~Roich, R.~Mokady, A.~H. Bermano, and D.~Cohen-Or, ``Pivotal tuning for
  latent-based editing of real images,'' \emph{ACM Transactions on graphics
  (TOG)}, vol.~42, no.~1, pp. 1--13, 2022.

\end{thebibliography}

\end{document}